\documentclass[10pt,twocolumn,letterpaper]{article}

\usepackage{cvpr}              

\usepackage{graphicx}
\usepackage{amsmath}
\usepackage{booktabs}
\usepackage{array}
\usepackage{multirow}
\usepackage{makecell}

\usepackage{colortbl}
\usepackage{pifont}

\usepackage{caption}
\usepackage{subcaption}

\definecolor{cvprblue}{rgb}{0.21,0.49,0.74}
\usepackage[pagebackref,breaklinks,colorlinks,allcolors=cvprblue]{hyperref}

\definecolor{lightgray}{gray}{0.9}

\def\paperID{7481} 
\def\confName{CVPR}
\def\confYear{2026}

\title{MSLoRA: Multi-Scale Low-Rank Adaptation via Attention Reweighting}

\author{Xu Yang \qquad Gady Agam\\
Department of Computer Science \\
Illinois Institute of Technology\\
{\tt\small xyang99@hawk.illinoistech.edu, agam@illinoistech.edu}
}

\begin{document}
\maketitle

\begin{abstract}
    We introduce MSLoRA, a backbone-agnostic, parameter-efficient adapter that reweights feature responses rather 
    than re-tuning the underlying backbone. Existing low-rank adaptation methods are mostly confined to vision 
    transformers (ViTs) and struggle to generalize across architectures. MSLoRA unifies adaptation for both convolutional neural networks (CNNs) and 
    ViTs by combining a low-rank linear projection with a multi-scale nonlinear transformation that jointly 
    modulates spatial and channel attention. The two components are fused through pointwise multiplication and 
    a residual connection, yielding a lightweight module that shifts feature attention while keeping pretrained 
    weights frozen. 
    Extensive experiments demonstrate that MSLoRA consistently improves transfer performance on classification, 
    detection, and segmentation tasks with roughly less than 5\% of backbone parameters. 
    The design further enables stable optimization, fast convergence, and strong cross-architecture 
    generalization. By reweighting rather than re-tuning, MSLoRA provides a simple and universal approach 
    for efficient adaptation of frozen vision backbones.
\end{abstract}

\setlength{\tabcolsep}{4pt}

\section{Introduction}
Deep learning~\cite{lecun2015deep} has driven remarkable progress in computer vision~\cite{taye2023theoretical}, 
largely through powerful pretrained models~\cite{deng2009imagenet,2023mmpretrain} that can be adapted to a wide 
range of downstream tasks~\cite{chen2019mmdetection}. Fine-tuning~\cite{liu2020deep} these pretrained networks 
has become the dominant paradigm, enabling strong performance with limited data and computation. 
As modern vision models continue to scale in size and capacity~\cite{he2016deep,xie2017aggregated,radosavovic2020designing,dosovitskiy2020image,liu2021swin}, 
the question of how to adapt them both efficiently and effectively~\cite{jia2022visual,hu2022lora} has become increasingly important.

Early approaches updated all model parameters for each new task. While flexible, full fine-tuning is computationally expensive and prone to overfitting when target datasets are small. Subsequent methods~\cite{hu2022lora,he2022parameter,houlsby2019parameter} strike a better balance between adaptation capacity and parameter efficiency, leading to partial fine-tuning~\cite{zaken2022bitfit,giannou2023expressive} and module-based adaptation strategies~\cite{chen2023sam,chen2022adaptformer,liu2022polyhistor}. These approaches demonstrate that pretrained representations can be reused effectively with minimal modification, often matching the performance of full fine-tuning~\cite{yin20231,yin2024parameter,sung2022vl}.

\begin{figure}[t]
    \centering
    \includegraphics[width=0.98\linewidth]{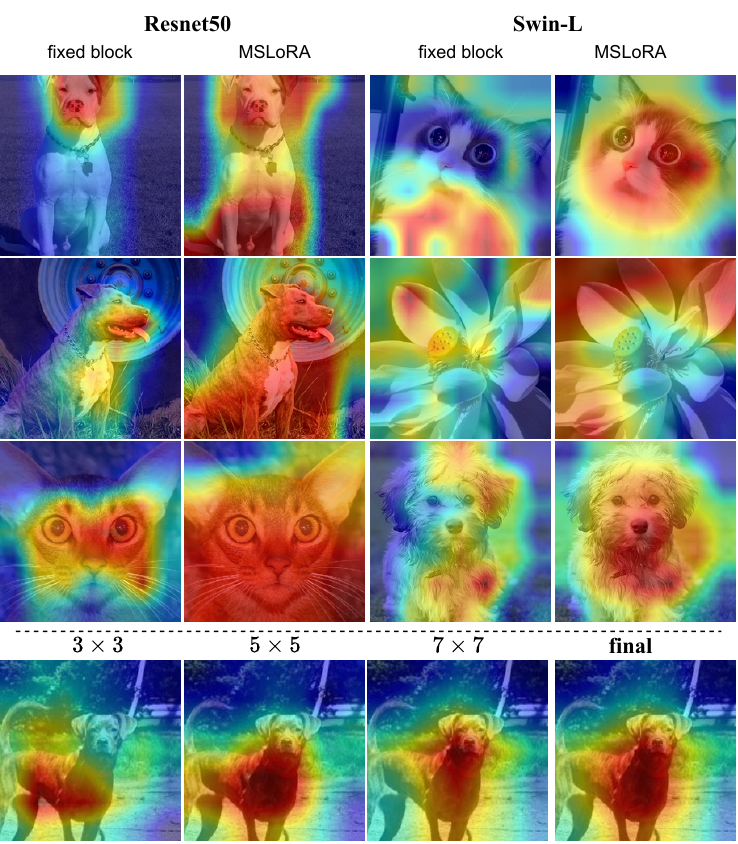}
    \caption{\textbf{Qualitative effects of attention reweighting.} \emph{Top:} A comparison between a fixed backbone block and our MSLoRA reweighting. Each MSLoRA layer uses fewer than 100K trainable parameters, whereas a fixed block typically exceeds 1M. \emph{Bottom:} MSLoRA focuses filters at different spatial scales to emphasize task-relevant features.}
    \label{fig:vis_example}
    \vspace{-1.5em}
\end{figure}

Despite this progress, current fine-tuning strategies~\cite{peters2019tune} face practical limitations. Fully fine-tuning large transformers or convolutional networks remains costly; backpropagation through randomly initialized task heads can distort useful pretrained semantics; and lightweight adapters can struggle to capture complex task-specific behavior or maintain training stability. Moreover, many low-rank adaptation methods~\cite{yin20231,liu2022p,ding2023parameter} closely follow LoRA and are primarily tailored to specific transformer backbones (e.g., Swin), often yielding suboptimal performance and limited generality. These challenges motivate a universal yet efficient approach that operates across both convolutional and transformer-based image backbones.

A key observation is that modern image backbones, whether CNNs or ViTs, share a multi-stage, multi-scale design that processes feature maps at progressively coarser resolutions~\cite{yang2022focal,he2016deep,liu2021swin,dosovitskiy2020image,radosavovic2020designing,liu2022swin,fan2021multiscale,li2022mvitv2}. This structural commonality suggests that a single adaptation module could be applied broadly across architectures. At the same time, pretraining on large datasets such as ImageNet~\cite{deng2009imagenet} induces semantics suited to classification, while downstream tasks such as detection and segmentation~\cite{lin2014microsoft,everingham2015pascal,zhou2017scene} require attention patterns over multiple objects and regions. Rather than relearning new features, a promising route is to preserve the useful pretrained semantics while reweighting spatial and channel-wise attention to suit the target task (Fig.~\ref{fig:vis_example}). Importantly, such a modulation module should be lightweight: the goal is to adjust the relative importance of existing features, not to replicate the role of a full backbone block. While LoRA~\cite{hu2022lora} effectively updates a few dominant directions in parameter space, 2D image features exhibit rich spatial structure that is not fully captured by purely linear projections. Combining linear and nonlinear components can therefore better support attention reweighting in two dimensions and stabilize the distribution of inputs to subsequent layers.

\begin{figure*}[t]
    \centering
    \includegraphics[width=1.0\textwidth]{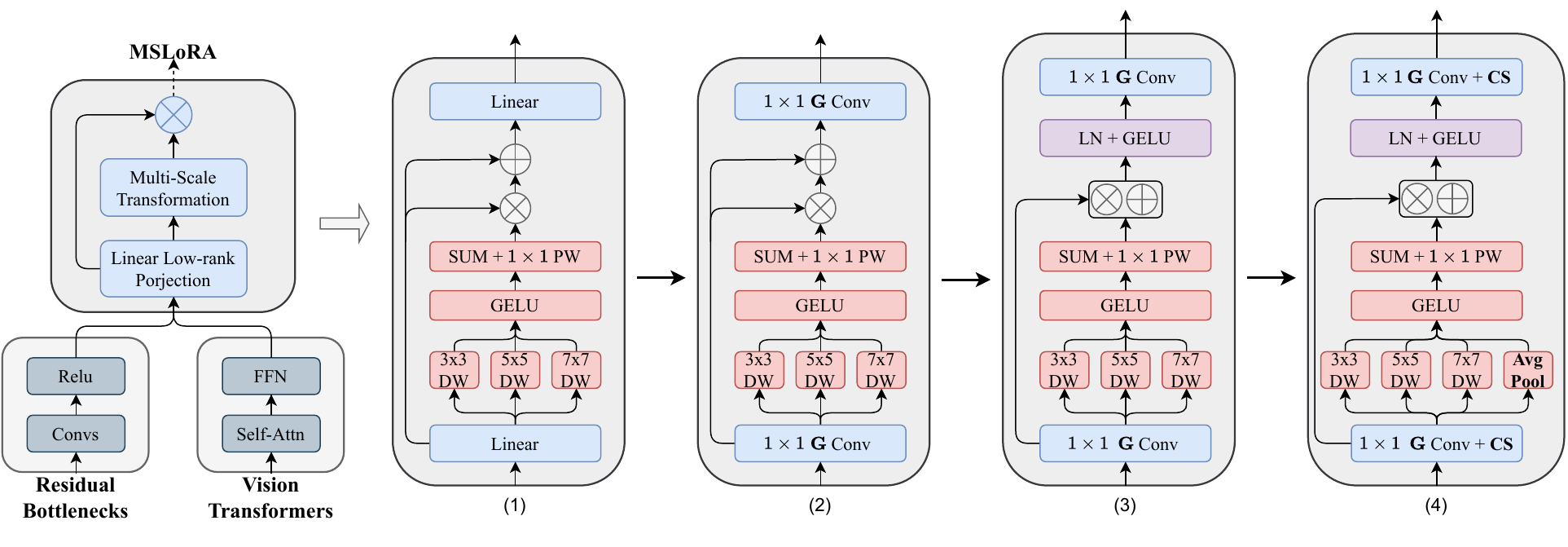}
    \caption{\textbf{MSLoRA overview.} The module (left) combines (i) a low-rank linear projection~\cite{hu2022lora} with (ii) a multi-scale nonlinear transformation~\cite{chansong2021impacts}. The two branches are fused via point-wise multiplication to produce an attention reweighting map. (1) shows the minimal design; (2) groups the linear projection to reduce parameters; (3) increases nonlinearity with additional activations; and (4) applies lightweight enhancements for further gains. We illustrate placement within both residual bottlenecks and transformer blocks (right) for generality.}
    \label{fig:module_design}
    \vspace{-1.5em}
\end{figure*}

The main contribution of this paper is 
\textbf{MSLoRA}, a simple and parameter-efficient module for universal attention reweighting. Let \(F\) denote the feature map from a frozen backbone. We adopt the residual form
\[
\hat{F} = F + \mathcal{H}(F),
\]
where \(\mathcal{H}(\cdot)\) is a lightweight learnable function that modulates the importance of spatial locations and channels. Our design targets two goals: (i) an efficient, architecture-agnostic reweighting function \(\mathcal{H}\) that applies to both CNN and transformer stages, and (ii) minimal trainable parameters~\cite{ding2023parameter,liu2022polyhistor}. Concretely, MSLoRA first applies a low-rank linear projection~\cite{hu2022lora}, implemented as a grouped \(1{\times}1\) convolution, to reduce dimensionality while preserving key directions, and then employs a multi-scale depth-wise convolutional stack~\cite{sandler2018mobilenetv2,zhang2018shufflenet,yang2022focal} followed by \textsc{GELU}~\cite{hendrycks2016gaussian} and a point-wise \(1{\times}1\) convolution~\cite{howard2017mobilenets,he2016deep} to capture diverse receptive fields. Lightweight nonlinear enhancements further stabilize optimization. We provide ablations on scale granularity, adapter rank, grouping, and enhancement choices, together with analysis of parameter and compute budgets.

Experimental results demonstrate the effectiveness of our approach.
On COCO~\cite{lin2014microsoft} with Cascade Mask R-CNN~\cite{cai2019cascade}, MSLoRA achieves 42.9 (+1.7) box mAP and 38.4 (+2.5) mask mAP on ResNet-50~\cite{he2016deep} with only 0.7M trainable parameters (2.7\%). On Swin-B~\cite{liu2022swin}, it reaches 52.7 (+0.3) box mAP and 45.9 (+0.8) mask mAP with 1.8M trainable parameters in the backbone (2.0\%). Additional results and analyses are presented below.

\section{Related Work}

Image backbones provide the foundational representations for modern vision systems~\cite{szegedy2015going,he2016deep}. Early convolutional networks established that hierarchical feature extraction~\cite{lin2017feature} generalizes across diverse tasks. Subsequent designs improved capacity via deeper architectures, residual connections, and multi-scale structures. More recently, ViTs~\cite{dosovitskiy2020image,liu2021swin,fan2021multiscale} shifted the focus to attention-based models that capture long-range dependencies and global context. In practice, both CNN and transformer backbones are typically pretrained on large datasets~\cite{deng2009imagenet} and then adapted to downstream problems through fine-tuning~\cite{kading2016fine,wang2022pre} or lightweight modifications.

Fine-tuning is the standard mechanism for adapting pretrained models to new tasks. Full fine-tuning~\cite{wang2022pre} updates all parameters and offers strong flexibility, but incurs substantial computational and memory cost. To improve efficiency, numerous strategies~\cite{zaken2022bitfit,giannou2023expressive} restrict updates to a subset of layers or parameters (e.g., classification heads or normalization statistics). Delta-tuning methods~\cite{chen2023sam,jia2022visual,he2023parameter,liu2022polyhistor} introduce small trainable modules that modify intermediate activations or weights while keeping the backbone largely frozen, striking a balance between efficiency and expressiveness. Recent work also explores dynamic or structured allocation of adaptation capacity as a function of task difficulty or layer importance.

Low-rank adaptation provides a compact mechanism for fine-tuning large models~\cite{si2024flora}. Rather than modifying the original weights directly, LoRA inserts low-rank matrices whose product approximates the task-specific update, substantially reducing the number of trainable parameters while preserving the representational power of the pretrained backbone. LoRA has been effective across modalities and architectures, including ViTs~\cite{dosovitskiy2020image}, enabling efficient adaptation with minimal computational overhead. Extensions explore rank scaling, task-specific parameterization, and combinations with other modular fine-tuning components.

Recent works have explored parameter-efficient fine-tuning (PEFT) in both vision and language domains. Xin et al.~\cite{xin2024parameter} provide a comprehensive survey on PEFT techniques for vision models, including low-rank methods like LoRA, highlighting their importance in reducing computational and memory costs while achieving competitive results. Furthermore, Sahay and Savakis~\cite{sahay2025mopeft} propose a mixture-of-PEFTs approach for the Segment Anything Model (SAM), demonstrating the benefits of combining multiple PEFT methods for improved segmentation tasks.

\section{Approach}

Modern vision systems typically rely on pretrained backbones that are later fine-tuned for specific tasks~\cite{peters2019tune}. We introduce a simple mechanism that exploits these pretrained features more effectively by reweighting activations instead of altering backbone parameters. Specifically, a lightweight module is inserted at the end of each block to modulate the importance of spatial locations and channels. Figure~\ref{fig:vis_example} illustrates the effect: a frozen ResNet~\cite{he2016deep} tends to emphasize only the most discriminative parts, while Swin~\cite{liu2022swin} can overlook crucial regions. After MSLoRA reweighting, attention spans the full object extent, improving fine-grained recognition and downstream detection/segmentation. Visualizations across different filter shapes further show that the resulting multi-scale behavior captures diverse spatial semantics.

Our goal is not to learn new features from scratch, but to reweight existing representations using a small number of parameters. Guided by this principle, we design a universal, architecture-agnostic module that applies to both CNN and transformer backbones (Fig.~\ref{fig:module_design}). MSLoRA comprises two components: (i) a linear projection branch and (ii) a nonlinear transformation branch. The combined reweighting is
\begin{equation}
\mathrm{MSLoRA}(F) = \mathrm{Proj}(F) \odot \mathrm{Trans}(F),
\end{equation}
where $\odot$ denotes element-wise multiplication and $F$ is the input feature map from the frozen backbone.

The projection branch $\mathrm{Proj}(F)$ follows a low-rank design akin to LoRA: we first down-project channels from $C_{\text{in}}$ to a fixed $C \ll C_{\text{in}}$, apply the nonlinear branch on this compact representation to reduce compute, and then up-project back to $C_{\text{in}}$ to interface with subsequent blocks. The nonlinear branch $\mathrm{Trans}(\cdot)$ produces a data-dependent modulation that captures spatial and channel-wise structure. We fuse the two branches by point-wise multiplication to obtain the final attention map and residual reweighting.

A full schematic is given in Figure~\ref{fig:module_design}. In the next sections we detail the linear projection (grouped $1{\times}1$ projections for parameter efficiency) and the multi-scale nonlinear transformation (depth-wise convolutions with different receptive fields, followed by \textsc{GELU} and a point-wise $1{\times}1$ conv), and we analyze the trade-offs among rank, grouping, and multi-scale design.

\begin{figure}[t]
    \centering
    \includegraphics[width=0.47\textwidth]{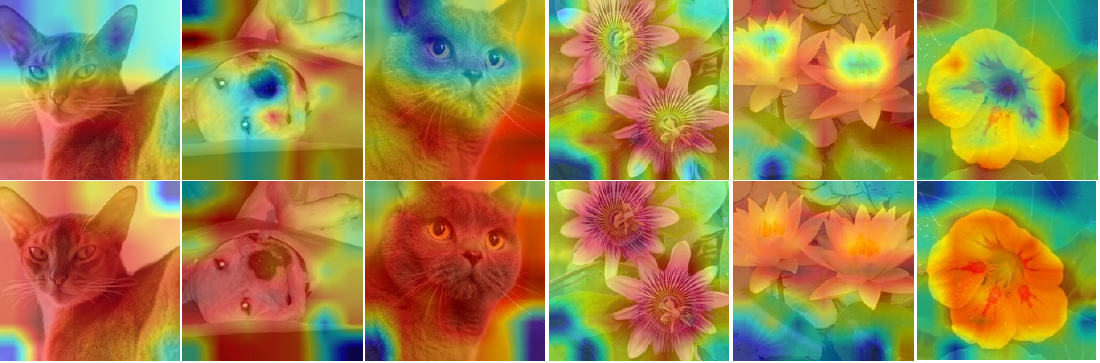}
    \caption{Comparison between standard fine-tuning (\emph{top}) and MSLoRA (\emph{bottom}). MSLoRA keeps the backbone frozen and learns lightweight reweighting for task-adaptive attention.}
    \label{fig:comp_finetune_mslora}
    \vspace{-1.5em}
\end{figure}

\subsection{Efficient Low-Rank Projection}
LoRA~\cite{hu2022lora} shows that linear attention weights can be adjusted effectively in a low-dimensional space. Following this insight, we hypothesize that the principal components of spatial semantics are also separable in a compact feature subspace. Directly applying a nonlinear transformation to high-dimensional features is computationally prohibitive, so we adopt a symmetric low-rank bottleneck: a down-projection to a smaller channel dimension, followed by an up-projection back to the original size. Our internal dimension is intentionally larger than that used in classic LoRA to preserve spatial detail while still reducing cost.

Concretely, we introduce a linear projection that reduces the channel dimension from $C_{\text{in}}$ to $C$ and place a symmetric layer at the end that expands $C$ back to $C_{\text{in}}$. We evaluate this design via an ablation over the low-rank size $C \in \{64,\,128,\,256,\,0.25\,C_{\text{in}}\}$ on an image classification benchmark~\cite{bossard2014food}.

We further analyze the parameter distribution within MSLoRA by defining 
\[
r \,=\, \frac{\text{proj.}}{\text{trans.}},
\]
the ratio of parameters in the linear projection branch to those in the nonlinear transformation branch. Table~\ref{tab:parameters_analysis} reports these counts. With a dense $1{\times}1$ projection (using $\mathbf{G}\!=\!1$ groups), the projection dominates the budget, especially on Swin-L, where $C_{\text{in}}$ is large. This is undesirable, since the nonlinear transformation is chiefly responsible for spatial reweighting. We posit that high-dimensional semantics are not uniformly distributed across channels, making a full-rank projection redundant.

To address this, we split the projection into $\mathbf{G}$ groups along the channel dimension, implemented as a grouped $1{\times}1$ convolution~\cite{xie2017aggregated,zhang2018shufflenet} (see Fig.~\ref{fig:module_design}). Grouping sharply reduces parameters in the projection branch and brings $r$ closer to $1$, better balancing capacity between the two branches. We ablate $\mathbf{G}\in\{1,2,4,8\}$ on both classification and downstream tasks.

\begin{table}[ht]
    \centering
    \small
    \begin{tabular}{c | c c c | c c c}
        \multirow{2}{*}{$\mathbf{G}$} & \multicolumn{3}{c}{Swin-L} & \multicolumn{3}{c}{ResNet-50} \\
                                       & proj. & trans. & ratio & proj. & trans. & ratio \\ \hline
        1                              & 6.9M  & 0.6M   & 17.2  & 1.4M  & 0.4M   & 3.50 \\
        2                              & 3.4M  & 0.6M   & 5.6   & 0.7M  & 0.4M   & 1.75 \\
        \rowcolor{lightgray}
        4                              & 1.7M  & 0.6M   & 2.8   & 0.3M  & 0.4M   & 0.75 \\
        8                              & 0.8M  & 0.6M   & 1.3   & 0.1M  & 0.4M   & 0.25 \\
        16                             & 0.4M  & 0.6M   & 0.6   & 0.1M  & 0.4M   & 0.25 \\
    \end{tabular}
    \caption{Parameter counts for different projection group sizes on ResNet-50 and Swin-L. \emph{proj.} and \emph{trans.} denote parameters in the projection and transformation branches, respectively; \emph{ratio} is their quotient. The \emph{trans.} column remains constant because the low-rank dimension is fixed.}
    \label{tab:parameters_analysis}
    \vspace{-1.5em}
\end{table}

\begin{table*}[h]
    \centering
    \setlength{\tabcolsep}{2pt}
    \small

    \begin{subtable}[t]{0.35\textwidth}
        \setlength{\tabcolsep}{1.5pt}
        \centering
        \begin{tabular}{@{}l r r c c r r c c@{}}
            $\mathbf{G}$ & \#p & \% & \small top-1 & \small top-5 & \#p & \% & \it bbox & \it mask \\
            \toprule
            ft. & 88M & 100\% & 92.7 & 98.6 & 23M & 92\% & 41.2 & 35.9 \\
            1   & 5.2M & 5.9\%  & 93.7 & 99.2 & 1.8M & 7.0\% & 42.5 & 38.1 \\
            2   & 2.9M & 3.2\%  & 92.2 & 99.1 & 1.1M & 4.2\% & 41.6 & 37.3 \\
            \rowcolor{lightgray}
            4   & 1.8M & 2.0\%  & 92.9 & 99.2 & 0.7M & 2.7\% & 41.9 & 37.8 \\
            8   & 1.2M & 1.3\%  & 92.6 & 99.1 & 0.6M & 2.3\% & 41.3 & 37.4 \\
        \end{tabular}
        \caption{\textbf{Projection group sizes.} We select $\mathbf{G}{=}4$ as a balanced trade-off between performance and trainable parameters.}
        \label{tab:proj_group_performance}
    \end{subtable}
    \hfill
    \begin{subtable}[t]{0.28\textwidth}
        \setlength{\tabcolsep}{2pt}
        \centering
        \begin{tabular}{@{}l r r c c@{}}
            dim. & \#p & \% & top-1 & top-5 \\
            \toprule
            64   & 0.8M & 0.9\% & 92.5 & 98.9 \\
            \rowcolor{lightgray}
            128  & 1.8M & 2.0\% & 92.9 & 99.2 \\
            256  & 4.3M & 4.8\% & 93.1 & 99.1 \\
            0.25 & 1.9M & 2.1\% & 92.9 & 99.2 \\
            \\
        \end{tabular}
        \caption{\textbf{Low-rank dimensions.} Accuracy improves with larger $C$, with diminishing returns beyond $128$.}
        \label{tab:low_rank_dim}
    \end{subtable}
    \hfill
    \begin{subtable}[t]{0.3\textwidth}
        \setlength{\tabcolsep}{2pt}
        \centering
        \begin{tabular}{@{}l r r c c@{}}
            filter sizes & \#p & \% & \it bbox & \it mask \\
            \toprule
            {[3]}         & 0.6M & 2.3\% & 41.1 & 37.2 \\
            {[3,5]}       & 0.6M & 2.3\% & 41.5 & 37.6 \\
            \rowcolor{lightgray}
            {[3,5,7]}     & 0.7M & 2.7\% & 41.9 & 37.8 \\
            {[3,5,7,9]}   & 0.9M & 3.4\% & 42.0 & 37.8 \\
            {[9,11,13]}   & 1.3M & 5.0\% & 41.9 & 37.8 \\
        \end{tabular}
        \caption{\textbf{Filter-size combinations.} Performance peaks at the \([3,5,7]\) multi-scale setting.}
        \label{tab:low_rank_filter_size}
    \end{subtable}

    \begin{subtable}[t]{0.35\textwidth}
        \setlength{\tabcolsep}{2pt}
        \centering
        \begin{tabular}{l r c c r c c}
            method    & \#p  & top-1 & top-5 & \#p  & \it bbox & \it mask \\
            \toprule
            linear    & 1.1M & 92.7  & 99.1  & 0.3M & 40.3 & 36.5 \\
            non-linear& 1.7M & 92.6  & 99.1  & 0.6M & 41.6 & 37.6 \\
            \rowcolor{lightgray}
            both      & 1.8M & 92.9  & 99.2  & 0.7M & 41.9 & 37.8 \\
        \end{tabular}
        \caption{\textbf{Reweighting variants.} Combining linear and nonlinear branches performs best across tasks.}
        \label{tab:reweight_method}
    \end{subtable}
    \hfill
    \begin{subtable}[t]{0.28\textwidth}
        \centering
        \begin{tabular}{l c c c c}
            method   & top-1 & top-5 & \it bbox & \it mask \\
            \toprule
            baseline & 92.9  & 99.2  & 41.9 & 37.8 \\
            \rowcolor{lightgray}
            enhance  & 93.0  & 99.1  & 42.5 & 38.2 \\
            \\
        \end{tabular}
        \caption{\textbf{Enhancing nonlinearity.} The enhanced variant consistently improves performance.}
        \label{tab:enhance_nonlinearity}
    \end{subtable}
    \hfill
    \begin{subtable}[t]{0.3\textwidth}
        \centering
        \setlength{\tabcolsep}{2pt}
        \begin{tabular}{l c c c c}
            method                 & top-1 & top-5 & \it bbox & \it mask \\
            \toprule
            $+$ avg. pooling       & 93.0  & 99.1  & 42.5 & 38.2 \\
            $+$ gate attn.         & 93.1  & 99.2  & 42.7 & 38.5 \\
            $+$ channel shuffle    & 93.2  & 99.2  & 42.9 & 38.4 \\
        \end{tabular}
        \caption{\textbf{Lightweight tricks.} Stacking simple tricks yields small but consistent gains.}
        \label{tab:enhance_trick}
    \end{subtable}

    \caption{\textbf{Overall ablations for MSLoRA.} $\mathbf{G}$ denotes the projection group size in the linear branch. \#p is the number of trainable parameters; \% is the fraction relative to the backbone. Classification results are on Food101 with Swin-B (Top-1/Top-5, \%), and detection/segmentation results are on COCO with ResNet-50 Cascade Mask R-CNN (bbox/mask mAP).}
    \vspace{-1.5em}
\end{table*}

\subsection{Multi-Scale Low-Rank Adaptation}
To capture spatial relationships in 2D features, we place a nonlinear transformation module after the linear projection. We use an independent projection for this branch (i.e., not shared with the linear branch) because the two serve different roles: the linear branch preserves the principal components in a compact space, while the nonlinear branch learns to recognize and differentiate spatial patterns.
The design draws inspiration from lightweight CNNs such as MobileNet~\cite{howard2017mobilenets,sandler2018mobilenetv2,zhang2018shufflenet}.
Unlike those primarily developed for classification, MSLoRA is also intended for downstream tasks such as detection and segmentation that demand richer spatial representations to handle dynamic input shapes and object scales.

To that end, we adopt multi-scale convolutions~\cite{yang2022focal} to provide diverse receptive fields (red boxes in Fig.~\ref{fig:module_design}). Concretely, we apply several parallel depth-wise convolution layers with different filter sizes and ablate combinations \([3], [3,5], [3,5,7], [3,5,7,9], [9,11,13]\) to identify a strong setting. Each branch is followed by a \textsc{GELU} activation, after which we sum the branches without normalization (the feature statistics are already stabilized by the preceding projection). A final point-wise convolution exchanges information across channels. Formally,
\begin{equation}
    \hat{F} = \mathrm{PW}\!\left(\,{\textstyle \sum\nolimits_k}\ \mathrm{GELU}\big(\mathrm{DW}_{k}(F)\big)\right)
\end{equation}
where $\mathrm{DW}_{k}$ and $\mathrm{PW}$ denote depth-wise and point-wise ($1{\times}1$) convolutions, respectively, and $k$ is the kernel size.
This transformation remains lightweight because both depth-wise and point-wise convolutions operate in the low-dimensional space. As reported in Table~\ref{tab:parameters_analysis}, the entire nonlinear branch uses fewer than 1M parameters even with a Swin-L backbone, yet plays a central role in reweighting spatial features.

\subsection{Reweighting Feature Attention}
We combine the linear projection and nonlinear transformation via element-wise multiplication to perform attention reweighting. For intuition, compare with self-attention~\cite{vaswani2017attention}:
{\small
\begin{equation*}
\begin{aligned}
  \mathrm{Attn}(Q,K,V) &= 
    \underbrace{\mathrm{Softmax}(QK^{\top})}_{\text{nonlinear}}\,
    \underbrace{V}_{\text{linear}}, \\[-1pt]
  \mathrm{MSLoRA}(F) &= 
    \underbrace{\mathrm{PW}\!\Big(\sum\nolimits_k
    \mathrm{GELU}\big(\mathrm{DW}_{k}(F)\big)\Big)}_{\text{nonlinear}}\,
    \underbrace{\mathrm{GConv}(F)}_{\text{linear}}
\end{aligned}
\end{equation*}
}
Considering Softmax as an activation, the structures are analogous: a nonlinear attention term modulates a linear value pathway. The key difference is computational: attention operates on token sequences with quadratic complexity in sequence length, which is prohibitive for a lightweight module on dense 2D maps. Moreover, global Softmax emphasizes all-token interactions, whereas spatial relationships in images are often local and multi-scale. We therefore replace global token attention with multi-scale convolutional attention followed by an activation to highlight salient patterns. The resulting region-aware attention map is multiplied with the projected features, and we add a residual connection for stable optimization 
(see $\otimes$ and $\oplus$ in Fig.~\ref{fig:module_design}, where $\otimes$ denotes element-wise multiplication $\odot$ between branches, and $\oplus$ indicates residual addition).

\subsection{Enhancing Nonlinearity}
Figure~\ref{fig:module_design} (part 2) shows the baseline module with a shortcut. While this improves performance, insufficient learning in the transformation branch can cause MSLoRA to degenerate toward a purely linear projection. To mitigate this, we modestly increase nonlinearity: after the multi-scale operations, we add a LayerNorm to further stabilize feature statistics, followed by an additional \textsc{GELU} to increase capacity (part 3 in Fig.~\ref{fig:module_design}). We ablate this enhancement on both classification and downstream tasks and observe consistent gains.

\subsection{Enhancement Strategies}
We explore several lightweight strategies to further improve MSLoRA.

\noindent\textbf{Global pooling.}
Standard convolutions have limited receptive fields, while high-level understanding for detection and segmentation benefits from global context~\cite{zhao2017pyramid,chen2014semantic}. Adding a global pooling branch summarizes image-level semantics and injects them back into local features, improving consistency and reducing ambiguity among similar local patterns.

\noindent\textbf{Gated attention.}
Inspired by attention mechanisms in CNNs~\cite{hu2018squeeze,graves2012long}, we add a lightweight gating to help the model emphasize informative responses, complementing our simple nonlinear branch. Instead of a full SE (Squeeze-and-Excitation) block which is less parameter-efficient, we learn a single attention map per parallel path and apply it multiplicatively, hence a gate that modulates feature strength with minimal overhead.

\noindent\textbf{Channel shuffle.}
Channel shuffle~\cite{zhang2018shufflenet} promotes information exchange across groups. Because our projection uses grouped $1{\times}1$ convolutions, shuffling can further mix channels between groups. Unlike ShuffleNet, MSLoRA is not strictly dependent on shuffling, given its inputs (from CNNs or ViTs) already arise from full-rank operations. Nevertheless, we still ablate channel shuffle as a common practice for grouped convolutions.

\begin{table}[ht]
    \centering
    \small
    \setlength{\tabcolsep}{2.5pt}
    \begin{tabular}{l l r r c c}
        Backbone & Method & \#p & \% & \multicolumn{1}{c}{AP$^{\text{bbox}}$} & \multicolumn{1}{c}{AP$^{\text{mask}}$} \\
        \toprule
        \multirow{5}{*}{ResNet-50}
            & fine-tune                         & 23.5M  & 92\%  & 41.2 & 35.9 \\
            & fixed                             & 0      & 0\%   & 14.0 & 12.7 \\
            & norm-tune~\cite{giannou2023expressive} & 45K    & $<$1\% & 29.2 & 26.0 \\
            & bias-tune~\cite{zaken2022bitfit}       & 23K    & $<$1\% & 35.8 & 31.7 \\
            & \textbf{\footnotesize MSLoRA}     & 0.7M   & 2.7\% & \textbf{42.9}{\scriptsize(+1.7)} & \textbf{38.4}{\scriptsize(+2.5)} \\
        \hline
        \multirow{2}{*}{ResNet-101}
            & fine-tune                         & 41.2M  & 95\%  & 42.9 & 37.3 \\
            & \textbf{\footnotesize MSLoRA}     & 1.6M   & 3.7\% & \textbf{44.0}{\scriptsize(+1.1)} & \textbf{39.2}{\scriptsize(+1.9)} \\
        \hline
        \multirow{2}{*}{ResNeXt-101}
            & fine-tune                         & 41.2M  & 95\%  & 44.3 & 38.3 \\
            & \textbf{\footnotesize MSLoRA}     & 1.6M   & 3.7\% & \textbf{44.8}{\scriptsize(+0.5)} & \textbf{39.9}{\scriptsize(+1.6)} \\
        \hline
        \multirow{2}{*}{RegNetX-4.0}
            & fine-tune                         & 22.2M  & 95\%  & 44.4 & 38.6 \\
            & \textbf{\footnotesize MSLoRA}     & 1.7M   & 7.6\% & 44.3{\scriptsize(-0.1)} & \textbf{39.7}{\scriptsize(+1.1)} \\
        \hline
        \multirow{7}{*}{Swin-Base}
            & fine-tune                         & 88M    & 100\% & 52.4 & 45.1 \\
            & LoRA~\cite{hu2022lora}            & 3.0M   & 3.4\% & 50.4 & 43.9 \\
            & AdaptFormer~\cite{chen2022adaptformer} & 1.6M & 1.7\% & 51.7 & 44.6 \\
            & LoRand~\cite{yin20231}            & 4.6M   & 5.2\% & 51.9 & 44.7 \\
            & Adapter~\cite{chen2022vision}     & 3.1M   & 3.5\% & 52.1 & 45.0 \\
            & \textbf{\footnotesize MSLoRA}     & 1.8M   & 2.0\% & \textbf{52.7}{\scriptsize(+0.3)} & \textbf{45.9}{\scriptsize(+0.8)} \\
            & \textbf{\footnotesize MSLoRA$^{\ast}$} & 5.2M & 5.9\% & \textbf{53.2}{\scriptsize(+0.8)} & \textbf{46.1}{\scriptsize(+1.0)} \\
        \bottomrule
    \end{tabular}
    \caption{COCO results with MSLoRA across backbones. \#p denotes trainable parameters; \% is relative to the backbone. MSLoRA$^{\ast}$ uses $\mathbf{G}{=}1$ in the projection branch. We bold results which outperform full fine-tune.}
    \label{tab:coco_performance}
    \vspace{-1.5em}
\end{table}

\subsection{Parameter Analysis}
We compute MSLoRA’s parameters with input channels $C_{\text{in}}$, low-rank width $D$, and $\mathbf{G}$ groups. The grouped $1{\times}1$ projections contribute $3C_{\text{in}}D/\mathbf{G}$ parameters; the multi-scale depth-wise convolutions (kernels $3,5,7$) add $(3^2\!+\!5^2\!+\!7^2)D = 83D$; and the point-wise $1{\times}1$ adds $D^2$. Thus resulting in a total parameter count of
\begin{equation}
P_{\text{MSLoRA}} = 
    \frac{3\,C_{\text{in}}D}{\mathbf{G}} + D^2 + 83D.
    \label{eq:total_params}
\end{equation}

\section{Experiments}

We comprehensively evaluate MSLoRA to assess its effectiveness, efficiency, and generalization across tasks and architectures.
Section \ref{sec:settings} outlines the default experimental settings, followed by ablations analyzing key design choices (Sec. \ref{sec:ablation}) and activation-level comparisons with full fine-tuning (Sec. \ref{sec:activation}).
We then present results on image classification (Sec. \ref{sec:classification}) and extend the evaluation to downstream detection and segmentation tasks (Sec. \ref{sec:downstream}).

\subsection{Default Settings}
\label{sec:settings}
We perform ablations by enabling one component at a time to isolate the contribution of each part of MSLoRA. 
Changes are non-cumulative unless otherwise specified. 
Unless stated, we set $\mathbf{G}{=}4$ and $D{=}128$.

We evaluate two representative settings for cross-validation:
(i) image classification on Swin-B using Food-101; and 
(ii) detection/segmentation on ResNet-50 using COCO with Cascade Mask R-CNN ($1{\times}$, 12 epochs). 
MSLoRA is trained with AdamW (weight decay $0.05$, $\beta_1{=}0.9$, $\beta_2{=}0.999$). 
The backbone is frozen, including BatchNorm~\cite{ioffe2015batch} statistics when present. 
DropPath~\cite{huang2016deep} remains active in the backbone, which slightly reduces overfitting in our experiments.

Training uses \texttt{RandomResizedCrop}(224) and horizontal flip ($p{=}0.5$); validation uses \texttt{Resize}(256) and \texttt{CenterCrop}(224). 
Ablation runs are trained for 50 epochs, and final comparisons for 100 epochs, with a batch size of 16. 
The initial learning rate is $3{\times}10^{-5}$, scheduled by cosine decay with a 20-epoch linear warmup.

For detection and segmentation, the initial learning rate is $2{\times}10^{-4}$ for CNN backbones and $1{\times}10^{-4}$ otherwise. 
All other hyperparameters follow the respective fine-tuning configurations. 
For pre-norm architectures, we prepend a LayerNorm~\cite{ba2016layer} at the MSLoRA input.

\begin{figure}[ht]
    \centering
    \includegraphics[width=0.5\textwidth]{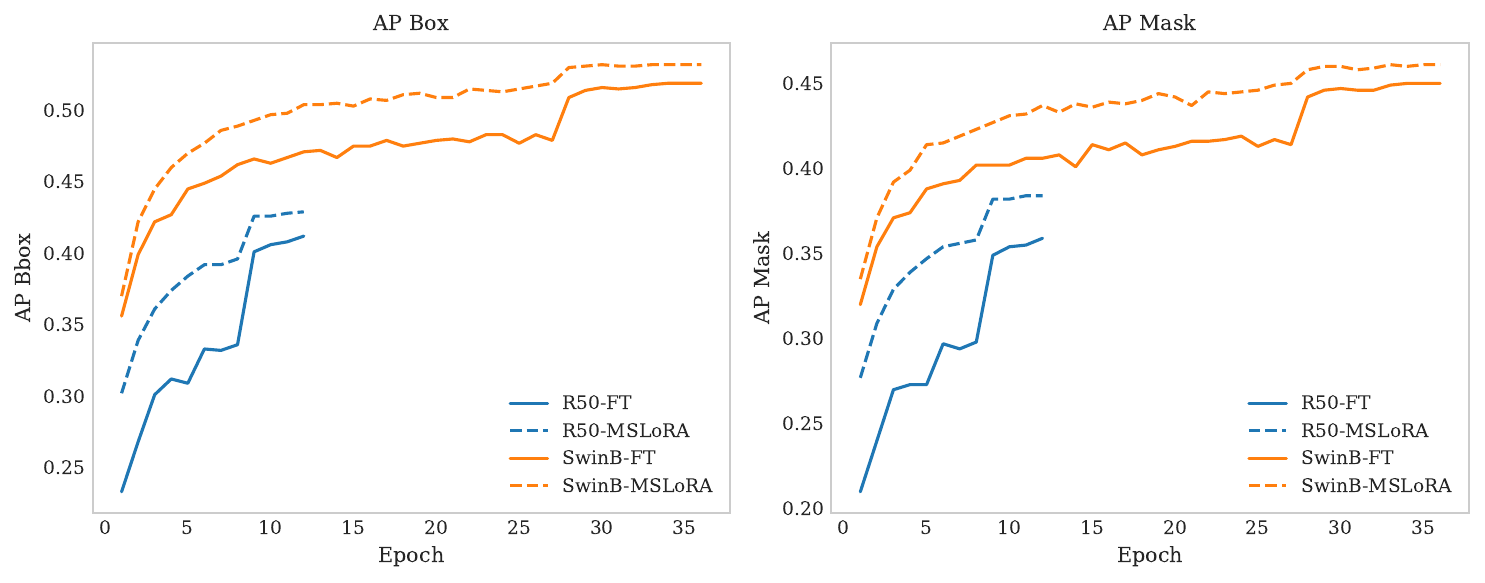}
    \caption{Comparison of full fine-tuning vs.\ MSLoRA across metrics.}
    \label{fig:metric_comparison}
    \vspace{-1.5em}
\end{figure}

\subsection{Ablation Study}
\label{sec:ablation}

We next analyze how individual design choices affect MSLoRA’s performance and efficiency. Each component is examined in isolation under the default configuration, revealing how projection grouping, multi-scale filtering, low-rank dimensionality, and reweighting strategies contribute to the model’s balance between adaptability and compactness.

\noindent\textbf{Projection group size.}
We vary the number of groups in the linear projection to balance parameters between the projection and transformation branches. Results are in Table~\ref{tab:proj_group_performance}. Performance peaks without grouping ($\mathbf{G}{=}1$), likely because information is preserved more completely through a dense projection; in this setting, both tasks surpass full fine-tuning: $93.7\%$ Top-1 on classification with only $3.8\%$ of the parameters, and $42.5$ AP$^{\text{bbox}}$ / $38.1$ AP$^{\text{mask}}$ on COCO with $7.0\%$ of the parameters. However, this choice yields the most unbalanced parameter split (Table~\ref{tab:parameters_analysis}): for Swin-L, the projection has $6.9$M parameters vs.\ $0.6$M in the nonlinear branch; ResNet-50 is better but still projection-dominated.

Introducing groups alleviates the imbalance and sharply reduces parameters, at a small accuracy cost. The worst case occurs at $\mathbf{G}{=}2$ as information loss in the projection is not yet fully compensated by the nonlinear branch. Increasing to $\mathbf{G}{=}4$ restores accuracy while retaining large savings: compared to $\mathbf{G}{=}1$, parameters drop by $\sim$60\% while maintaining strong performance ($92.9\%$ Top-1 with $2.3$M parameters; $41.9$ AP$^{\text{bbox}}$ / $37.8$ AP$^{\text{mask}}$ with $0.7$M parameters). We therefore adopt $\mathbf{G}{=}4$ as the default in subsequent experiments. Beyond this (e.g., $\mathbf{G}{=}8$), performance degrades noticeably and we do not consider larger group sizes further.

\begin{table}[ht]
    \centering
    \small
    \begin{tabular}{l | l c c}
        \multirow{2}{*}{Backbone} & \multirow{2}{*}{Method} & \textbf{Pascal VOC} & \textbf{ADE20K} \\
                                   &                          & (RetinaNet)          & (UPerNet)       \\
        \hline
        \multirow{2}{*}{ResNet-50}  & fine-tune                & 77.26 & 40.70 \\
                                    & \textbf{MSLoRA}          & \textbf{77.83}{\scriptsize(+0.57)} & \textbf{41.62}{\scriptsize(+0.92)} \\
        \hline
        \multirow{2}{*}{ResNet-101} & fine-tune                & 78.10 & 42.91 \\
                                    & \textbf{MSLoRA}          & \textbf{78.82}{\scriptsize(+0.72)} & \textbf{43.06}{\scriptsize(+0.15)} \\
        \hline
        \multirow{2}{*}{Swin-L}     & fine-tune                & 83.40 & 50.13 \\
                                    & \textbf{MSLoRA}          & \textbf{84.49}{\scriptsize(+1.09)} & \textbf{50.38}{\scriptsize(+0.25)} \\
    \end{tabular}
    \caption{Comparison on Pascal VOC and ADE20K.}
    \label{tab:other_performance}
    \vspace{-1.5em}
\end{table}

\noindent\textbf{Multi-scale filter sizes.}
We ablate the filter-size combinations used in the nonlinear branch. Because classification on small inputs is relatively insensitive to receptive field, we report COCO results only (Table~\ref{tab:low_rank_filter_size}). Using a single $3{\times}3$ filter performs worst, dropping $0.8$ AP$^{\text{bbox}}$ and $0.6$ AP$^{\text{mask}}$ versus the default. As input resolution grows, the patterns we aim to reweight cannot be fully captured by small receptive fields. Adding a $5{\times}5$ branch yields a further $+0.4$ AP for both bbox and mask, and introducing a $7{\times}7$ branch reaches the best trade-off (41.9 AP$^{\text{bbox}}$, 37.8 AP$^{\text{mask}}$). Multi-scale filters help capture objects at diverse sizes within the same image, improving adaptation. Larger kernels such as $9{\times}9$ provide no additional gains, suggesting capacity is saturated. Since the parameter increase remains modest with depth-wise convolutions, we adopt $(3,5,7)$ as the default. Very large sets~\cite{yang2022focal} like $(9,11,13)$ similarly show no improvement.

\noindent\textbf{Low-rank dimension.}
We ablate fixed low-rank sizes versus a depth-adaptive choice (e.g., $0.25C$). Although an adaptive rule seems natural as channels grow with depth, it introduces two challenges: (i) early stages receive very small widths (e.g., $8$–$16$), which weakens spatial modeling and leads to information loss during reweighting; (ii) later stages become excessively wide. Since the dominant term in MSLoRA’s parameter count is $D^2$ (Eq.~\ref{eq:total_params}), doubling the channel dimension can quadruple the number of parameters. This is undesirable for parameter-efficient fine-tuning (PEFT), where compactness is critical.

We therefore prefer a fixed $D$. As shown in Table~\ref{tab:low_rank_dim}, $D{=}64$ uses only $0.9\%$ of the parameters but drops to $92.5\%$ Top-1 accuracy. Increasing to $D{=}128$ offers a favorable trade-off ($92.9\%$ Top-1 with $2.0\%$ parameters), while $D{=}256$ provides only marginal gains ($93.1\%$ Top-1) at more than double the parameter cost ($4.8\%$). The relative option $0.25C$ achieves similar accuracy to $D{=}128$ but with slightly higher overhead ($2.1\%$ parameters). We adopt $D{=}128$ as the default, as smaller dimensions underfit in later layers and larger ones over-parameterize earlier layers. More sophisticated adaptive schemes are possible, but we keep the design simple for clarity and efficiency.

\begin{table*}[ht]
    \centering
    \begin{tabular}{l | l r r  c c  c c  c c  c c}
        \multirow{2}{*}{Backbone}  & \multirow{2}{*}{Method} & \multirow{2}{*}{\#p} & \multirow{2}{*}{\%} & \multicolumn{2}{c}{\textbf{Oxford Pets}~\cite{parkhi2012cats}} & \multicolumn{2}{c}{\textbf{Flower102}~\cite{nilsback2008automated}} & \multicolumn{2}{c}{\textbf{Food101}~\cite{bossard2014food}} & \multicolumn{2}{c}{\textbf{Average}} \\
                                   &                          &                      &                     & top-1       & top-5                      & top-1        & top-5                   & top-1         & top-5                & top-1         & top-5       \\
        \hline
        \multirow{2}{*}{Swin-B}    & fine-tune                &  88M                 &    100\%            & 94.0038     & 99.5639                    & 99.5772      & 99.9187                 & 92.7683       & 98.6772              & 95.4497       & 99.3866      \\                       
                                   & \textbf{\footnotesize MSLoRA} &  1.8M                &    2.0\%            & \textbf{94.9305}     & \textbf{99.7820}  & \textbf{99.7235}   & \textbf{99.9675}  & \textbf{93.3743} & \textbf{99.2515}  & \textbf{96.0094} &\textbf{99.6670}  \\
        \hline
        \multirow{5}{*}{Swin-L}    & fine-tune                &  197M                &    100\%            & 94.7397     & 99.6729                    & 99.6910      & 99.9512                 & 93.8456       & 99.0852              & 96.0921       & 99.5697      \\                                                                                                                                                                                                                    
                                   & LoRA                     &  4.5M                &    2.2\%            & 95.1485     & 99.8910                    & 99.5446      & 99.8536                 & 93.1960       & 99.0059              & 95.9630       & 99.5835     \\                                                                                                                                                                                                                    
                                   & Adapter                  &  4.6M                &    2.3\%            & 95.3393     & 99.8092                    & 99.5934      & 99.8536                 & 93.2356       & 99.0297              & 96.0561       & 99.5641      \\                                                                                                                                                                                                                    
                                   & LoRand                   &  5.2M                &    2.6\%            & 95.3515     & 99.8910                    & 99.5725      & 99.8536                 & 93.2753       & 99.0059              & 96.0664       & 99.5835      \\                                                                                                                                                                                                                    
                                   & \textbf{\footnotesize MSLoRA}         &  2.3M                &    1.4\%            & \textbf{95.5574}     & \textbf{99.8637}  & \textbf{99.7235}   & \textbf{99.9675}  & \textbf{94.1941} & \textbf{99.3505}  & \textbf{96.4916} & \textbf{99.7272}             \\                                                                                                                                                                                                                  
    \end{tabular}
    \caption{Comparison with Swin Transformer on image classification tasks.}
    \label{tab:classification_performance}
    \vspace{-1.5em}
\end{table*}

\noindent\textbf{Reweighting method.}
We evaluate the effectiveness of our reweighting design by adding a nonlinear branch and combining it with the linear projection via point-wise multiplication (Table~\ref{tab:reweight_method}). 

For classification, using only the linear projection (92.7\% Top-1) slightly outperforms using only the nonlinear branch (92.6\%). This mirrors observations in NLP: after large-scale pretraining, representations already encode rich nonlinear dependencies, so adapting to a new classification task often amounts to redistributing emphasis over existing subspaces. In this setting, a linear projection can be sufficient. Still, combining linear and nonlinear branches yields a modest gain (92.9\%), as the nonlinear path captures residual spatial patterns that the linear path cannot.

For detection/segmentation, the trend reverses. These tasks demand precise localization and model interactions among multiple objects and contexts, which linear transforms alone cannot fully capture. Here, the nonlinear branch is crucial: linear-only performs worst (40.3 AP$^{\text{bbox}}$, 36.5 AP$^{\text{mask}}$), nonlinear-only is stronger (41.6 / 37.6), and the combined design performs best (41.9 / 37.8).

\noindent\textbf{Enhancing nonlinearity.}
To prevent degeneration toward a purely linear projection, we add a normalization layer and an activation after the multi-scale block. As shown in Table~\ref{tab:enhance_nonlinearity}, this yields consistent gains on both tasks: for classification, Top-1 improves to 93.0\% (+0.1), surpassing baseline sightly; for downstream detection/segmentation, results rise to 42.5 AP$^{\text{bbox}}$ (+0.6) and 38.2 AP$^{\text{mask}}$ (+0.4). These improvements support our claim that additional nonlinearity increases the module’s capacity to capture complex spatial relationships while remaining lightweight.

\noindent\textbf{Additional enhancement strategies.}
Although each of the three techniques has proven effective in prior work, applying them \emph{individually} to MSLoRA yields little benefit. We attribute this to their complementary but incomplete roles: global pooling injects holistic context but can over-smooth fine details; gated attention adaptively emphasizes salient responses but needs sufficiently diverse cues; channel shuffle promotes cross-group exchange yet does not by itself improve contextual reasoning. Stacking them progressively compensates for these limitations. Global pooling provides a scene-level prior, gating sharpens relevance, and shuffling strengthens inter-channel communication. As shown in Table~\ref{tab:enhance_trick}, this synergy produces consistent gains, suggesting the joint integration yields a more balanced and expressive representation for both classification and downstream tasks.

\subsection{Activation-level Comparison with Fine-tuning}
\label{sec:activation}

We analyze the behavioral differences between full fine-tuning and MSLoRA’s reweighting. 
Figure~\ref{fig:comp_finetune_mslora} shows that MSLoRA tends to recover more complete object extents, whereas full fine-tuning often concentrates on highly discriminative parts. 
We hypothesize that reweighting imposes a useful regularization by constraining the update space, encouraging the model to favor more generalizable features. 
In contrast, unconstrained parameter updates in full fine-tuning can overfit to task-specific patterns. 
As shown in Figure~\ref{fig:metric_comparison}, MSLoRA converges faster and achieves higher final accuracy on COCO.

\subsection{Experiments on Image Classification}
\label{sec:classification}

We evaluate MSLoRA on three datasets: Oxford Pets~\cite{parkhi2012cats}, Flower102~\cite{nilsback2008automated}, and Food101~\cite{bossard2014food}. \emph{Pets} (37 classes, $\sim$7k images) is close to ImageNet and tests domain adaptation to similar categories. \emph{Flower102} (102 classes, $\sim$8k images) exhibits large scale/pose/illumination variation and probes fine-grained discrimination. \emph{Food101} (101 classes, 101k images) contains more complex scenes and categories that largely do not appear in ImageNet~\cite{deng2009imagenet}, stressing low-level cue aggregation and composition; we also use it for ablations due to its difficulty. We train for 100 epochs with an initial learning rate of $3\times10^{-5}$ (cosine decay, 20-epoch warmup). When using MSLoRA, the backbone is frozen and the enhancement tricks are disabled here (not helpful for these datasets). As summarized in Table~\ref{tab:classification_performance}, MSLoRA consistently outperforms full fine-tuning and other PEFT baselines across all datasets.

\subsection{Experiments on Downstream Tasks}
\label{sec:downstream}

To assess generalization beyond classification, we evaluate MSLoRA on object detection and segmentation benchmarks: COCO~\cite{lin2014microsoft} (detection and instance segmentation), Pascal VOC~\cite{everingham2015pascal} (detection), and ADE20K~\cite{zhou2017scene} (semantic segmentation). We use diverse backbones including ResNet, ResNeXt, RegNet, and Swin to test robustness across architectures.

\noindent\textbf{Results on COCO.}
We first evaluate MSLoRA with Cascade Mask R-CNN~\cite{cai2019cascade}. CNN backbones are trained for $1{\times}$ (12 epochs) and Swin for $3{\times}$ (36 epochs).  
As shown in Table~\ref{tab:coco_performance}, MSLoRA consistently surpasses full fine-tuning while requiring only a small fraction of trainable parameters.  
With ResNet-50, it achieves 42.9 (+1.7) AP$^{\text{bbox}}$ and 38.4 (+2.5) AP$^{\text{mask}}$ using just 0.7M (2.7\%) parameters.  
On deeper CNNs, the gains remain consistent: ResNet-101 reaches 44.0 (+1.1) / 39.2 (+1.9), and ResNeXt-101 achieves 44.8 (+0.5) / 39.9 (+1.6).  
For RegNetX, MSLoRA matches AP$^{\text{bbox}}$ (44.3 vs.\ 44.4) while improving AP$^{\text{mask}}$ (39.7 vs.\ 38.6).  
With Swin-Base, it attains 52.8 (+0.4) AP$^{\text{bbox}}$ and 45.5 (+0.4) AP$^{\text{mask}}$ using only 1.8M (2.0\%) parameters.  
Because some PEFT baselines underperform full fine-tuning under identical settings, we also report a higher-capacity variant with $\mathbf{G}{=}1$ (5.2M; 5.9\%), which further improves to 53.2 (+0.8) / 46.1 (+1.0), outperforming full fine-tuning by a clear margin.

\noindent\textbf{Results on other benchmarks.}
We next evaluate MSLoRA on Pascal VOC and ADE20K to examine generalization on smaller datasets.  
We use RetinaNet~\cite{lin2017focal} for Pascal VOC and UPerNet~\cite{xiao2018unified} for ADE20K, training all models for $1{\times}$.  
For RetinaNet, the learning rate is $10^{-4}$ for all methods.  
For UPerNet, we use $2\times10^{-4}$ for ResNet and $1.2\times10^{-4}$ for Swin, with weight decay $0.1$.  
Table~\ref{tab:other_performance} summarizes the results.

Across all benchmarks, MSLoRA consistently outperforms full fine-tuning while using far fewer parameters.  
With ResNet-50, it achieves 77.83 (+0.57) mAP on Pascal VOC and 41.62 (+0.92) mIoU on ADE20K.  
On ResNet-101, performance rises to 78.82 (+0.72) mAP and 43.06 (+0.15) mIoU.  
With Swin-L, MSLoRA reaches 84.49 (+1.09) mAP and 50.38 (+0.25) mIoU.  
These improvements across datasets and architectures demonstrate MSLoRA’s strong adaptability and generalization to diverse downstream tasks.


\section{Conclusion}
Pretrain–then–finetune remains the dominant paradigm in vision, yet full parameter updates are often inefficient and susceptible to overfitting. We revisited how pretrained representations can be more effectively adapted and proposed MSLoRA, a lightweight, parameter-efficient reweighting module that adapts features through multi-scale low-rank transformations.

Although MSLoRA aims for universality across vision backbones, its relative gains on Swin Transformer appear more modest compared to CNNs. We hypothesize this stems from two factors: (i) Swin's built-in robustness against overfitting through mechanisms like LayerNorm~\cite{ba2016layer} and DropPath~\cite{huang2016deep}; and (ii) the fundamental difference in how self-attention~\cite{vaswani2017attention} captures representations versus CNNs. Consequently, MSLoRA's CNN-based reweighting may not fully align with the complex dependencies in self-attention, leading to smaller improvements. Nonetheless, MSLoRA still exceeds full fine-tuning on Swin under default settings and shows further gains with increased parameter budgets. Future work could incorporate a lightweight attention-based transformation branch to better complement transformer features.

Through extensive ablations, we analyzed how multi-scale filtering, attention reweighting, and low-rank adaptation jointly enable strong accuracy–efficiency trade-offs. Experiments across diverse backbones (CNNs and Transformers) and downstream tasks—including classification, detection, and segmentation—demonstrate that MSLoRA consistently matches or surpasses full fine-tuning while updating only a small fraction of parameters.

By showing that careful reweighting of pretrained features can rival full optimization, this work highlights a new direction for scalable and general adaptation. We hope MSLoRA encourages further research into modular, efficient, and interpretable approaches for tuning large vision models.

{
  \small
  \bibliographystyle{ieeenat_fullname}
  \bibliography{mslora}

@inproceedings{he2016deep,
  title={Deep residual learning for image recognition},
  author={He, Kaiming and Zhang, Xiangyu and Ren, Shaoqing and Sun, Jian},
  booktitle={Proceedings of the IEEE conference on computer vision and pattern recognition},
  pages={770--778},
  year={2016}
}

@inproceedings{xie2017aggregated,
  title={Aggregated residual transformations for deep neural networks},
  author={Xie, Saining and Girshick, Ross and Doll{\'a}r, Piotr and Tu, Zhuowen and He, Kaiming},
  booktitle={Proceedings of the IEEE conference on computer vision and pattern recognition},
  pages={1492--1500},
  year={2017}
}

@inproceedings{radosavovic2020designing,
  title={Designing network design spaces},
  author={Radosavovic, Ilija and Kosaraju, Raj Prateek and Girshick, Ross and He, Kaiming and Doll{\'a}r, Piotr},
  booktitle={Proceedings of the IEEE/CVF conference on computer vision and pattern recognition},
  pages={10428--10436},
  year={2020}
}

@article{hu2022lora,
  title={LoRA: Low-Rank Adaptation of Large Language Models},
  author={Hu, Edward J and Shen, Yelong and Wallis, Phillip and Allen-Zhu, Zeyuan and Li, Yuanzhi and Wang, Shean and Wang, Lu and Chen, Weizhu},
  journal={International Conference on Learning Representations},
  year={2022},
  OPTurl={https://openreview.net/forum?id=nZeVKeeFYf9}
}

@article{he2022parameter,
  title={Parameter-efficient fine-tuning for vision transformers},
  author={He, Xuehai and Li, Chunyuan and Zhang, Pengchuan and Yang, Jianwei and Wang, Xin Eric},
  journal={arXiv preprint arXiv:2203.16329},
  year={2022}
}

@inproceedings{houlsby2019parameter,
  title={Parameter-efficient transfer learning for {NLP}},
  author={Houlsby, Neil and Giurgiu, Andrei and Jastrzebski, Stanislaw and Morrone, Bruna and De Laroussilhe, Quentin and Gesmundo, Andrea and Attariyan, Mona and Gelly, Sylvain},
  booktitle={International conference on machine learning},
  pages={2790--2799},
  year={2019},
  organization={PMLR}
}

@inproceedings{zaken2022bitfit,
  title={Bitfit: Simple parameter-efficient fine-tuning for transformer-based masked language-models},
  author={Zaken, Elad Ben and Goldberg, Yoav and Ravfogel, Shauli},
  booktitle={Proceedings of the 60th Annual Meeting of the Association for Computational Linguistics (Volume 2: Short Papers)},
  pages={1--9},
  year={2022}
}

@article{giannou2023expressive,
  title={The expressive power of tuning only the Norm layers},
  author={Giannou, Angeliki and Rajput, Shashank and Papailiopoulos, Dimitris},
  journal={arXiv preprint arXiv:2302.07937},
  volume={2},
  number={4},
  year={2023}
}

@inproceedings{jia2022visual,
  title={Visual prompt tuning},
  author={Jia, Menglin and Tang, Luming and Chen, Bor-Chun and Cardie, Claire and Belongie, Serge and Hariharan, Bharath and Lim, Ser-Nam},
  booktitle={European conference on computer vision},
  pages={709--727},
  year={2022},
  organization={Springer}
}

@inproceedings{chen2023sam,
  title={SAM-Adapter: Adapting Segment Anything in Underperformed Scenes},
  author={Chen, Tianrun and Zhu, Lanyun and Deng, Chaotao and Cao, Runlong and Wang, Yan and Zhang, Shangzhan and Li, Zejian and Sun, Lingyun and Zang, Ying and Mao, Papa},
  booktitle={Proceedings of the IEEE/CVF International Conference on Computer Vision Workshops},
  pages={3367--3375},
  year={2023}
}

@article{chen2022adaptformer,
  title={Adaptformer: Adapting vision transformers for scalable visual recognition},
  author={Chen, Shoufa and Ge, Chongjian and Tong, Zhan and Wang, Jiangliu and Song, Yibing and Wang, Jue and Luo, Ping},
  journal={Advances in Neural Information Processing Systems},
  volume={35},
  pages={16664--16678},
  year={2022}
}

@article{chen2022vision,
  title={Vision transformer adapter for dense predictions},
  author={Chen, Zhe and Duan, Yuchen and Wang, Wenhai and He, Junjun and Lu, Tong and Dai, Jifeng and Qiao, Yu},
  journal={arXiv preprint arXiv:2205.08534},
  year={2022}
}

@inproceedings{he2023parameter,
  title={Parameter-efficient model adaptation for vision transformers},
  author={He, Xuehai and Li, Chunyuan and Zhang, Pengchuan and Yang, Jianwei and Wang, Xin Eric},
  booktitle={Proceedings of the AAAI Conference on Artificial Intelligence},
  volume={37},
  number={1},
  pages={817--825},
  year={2023}
}

@inproceedings{zhang2018shufflenet,
  title={Shufflenet: An extremely efficient convolutional neural network for mobile devices},
  author={Zhang, Xiangyu and Zhou, Xinyu and Lin, Mengxiao and Sun, Jian},
  booktitle={Proceedings of the IEEE conference on computer vision and pattern recognition},
  pages={6848--6856},
  year={2018}
}

@inproceedings{lin2017focal,
  title={Focal loss for dense object detection},
  author={Lin, Tsung-Yi and Goyal, Priya and Girshick, Ross and He, Kaiming and Doll{\'a}r, Piotr},
  booktitle={Proceedings of the IEEE international conference on computer vision},
  pages={2980--2988},
  year={2017}
}

@article{cai2019cascade,
  title={Cascade {R-CNN}: High quality object detection and instance segmentation},
  author={Cai, Zhaowei and Vasconcelos, Nuno},
  journal={IEEE transactions on pattern analysis and machine intelligence},
  volume={43},
  number={5},
  pages={1483--1498},
  year={2019},
  publisher={IEEE}
}

@article{lecun2015deep,
  title={Deep learning},
  author={LeCun, Yann and Bengio, Yoshua and Hinton, Geoffrey},
  journal={nature},
  volume={521},
  number={7553},
  pages={436--444},
  year={2015},
  publisher={Nature Publishing Group UK London}
}

@article{chen2019mmdetection,
  title={MMDetection: Open mmlab detection toolbox and benchmark},
  author={Chen, Kai and Wang, Jiaqi and Pang, Jiangmiao and Cao, Yuhang and Xiong, Yu and Li, Xiaoxiao and Sun, Shuyang and Feng, Wansen and Liu, Ziwei and Xu, Jiarui and others},
  journal={arXiv preprint arXiv:1906.07155},
  year={2019}
}

@misc{2023mmpretrain,
    title={OpenMMLab's Pre-training Toolbox and Benchmark},
    author={MMPreTrain Contributors},
    howpublished = {\url{https://github.com/open-mmlab/mmpretrain}},
    year={2023}
}

@article{liu2022polyhistor,
  title={Polyhistor: Parameter-efficient multi-task adaptation for dense vision tasks},
  author={Liu, Yen-Cheng and Ma, Chih-Yao and Tian, Junjiao and He, Zijian and Kira, Zsolt},
  journal={Advances in Neural Information Processing Systems},
  volume={35},
  pages={36889--36901},
  year={2022}
}

@inproceedings{deng2009imagenet,
  title={Imagenet: A large-scale hierarchical image database},
  author={Deng, Jia and Dong, Wei and Socher, Richard and Li, Li-Jia and Li, Kai and Fei-Fei, Li},
  booktitle={2009 IEEE conference on computer vision and pattern recognition},
  pages={248--255},
  year={2009},
  organization={Ieee}
}

@article{liu2020deep,
  title={Deep learning for generic object detection: A survey},
  author={Liu, Li and Ouyang, Wanli and Wang, Xiaogang and Fieguth, Paul and Chen, Jie and Liu, Xinwang and Pietik{\"a}inen, Matti},
  journal={International journal of computer vision},
  volume={128},
  number={2},
  pages={261--318},
  year={2020},
  publisher={Springer}
}

@article{dosovitskiy2020image,
  title={An image is worth 16x16 words: Transformers for image recognition at scale},
  author={Dosovitskiy, Alexey and Beyer, Lucas and Kolesnikov, Alexander and Weissenborn, Dirk and Zhai, Xiaohua and Unterthiner, Thomas and Dehghani, Mostafa and Minderer, Matthias and Heigold, Georg and Gelly, Sylvain and Uszkoreit, Jakob and Houlsby, Neil},
  journal={arXiv preprint arXiv:2010.11929},
  year={2020}
}

@inproceedings{liu2021swin,
  title={Swin transformer: Hierarchical vision transformer using shifted windows},
  author={Liu, Ze and Lin, Yutong and Cao, Yue and Hu, Han and Wei, Yixuan and Zhang, Zheng and Lin, Stephen and Guo, Baining},
  booktitle={Proceedings of the IEEE/CVF international conference on computer vision},
  pages={10012--10022},
  year={2021}
}

@article{taye2023theoretical,
  title={Theoretical understanding of convolutional neural network: Concepts, architectures, applications, future directions},
  author={Taye, Mohammad Mustafa},
  journal={Computation},
  volume={11},
  number={3},
  pages={52},
  year={2023},
  publisher={MDPI}
}

@inproceedings{chansong2021impacts,
  title={Impacts of kernel size on different resized images in object recognition based on convolutional neural network},
  author={Chansong, Danupon and Supratid, Siriporn},
  booktitle={2021 9th international electrical engineering congress (iEECON)},
  pages={448--451},
  year={2021},
  organization={IEEE}
}

@article{ding2023parameter,
  title={Parameter-efficient fine-tuning of large-scale pre-trained language models},
  author={Ding, Ning and Qin, Yujia and Yang, Guang and Wei, Fuchao and Yang, Zonghan and Su, Yusheng and Hu, Shengding and Chen, Yulin and Chan, Chi-Min and Chen, Weize and others},
  journal={Nature machine intelligence},
  volume={5},
  number={3},
  pages={220--235},
  year={2023},
  publisher={Nature Publishing Group UK London}
}

@article{everingham2015pascal,
  title={The pascal visual object classes challenge: A retrospective},
  author={Everingham, Mark and Eslami, SM Ali and Van Gool, Luc and Williams, Christopher KI and Winn, John and Zisserman, Andrew},
  journal={International journal of computer vision},
  volume={111},
  number={1},
  pages={98--136},
  year={2015},
  publisher={Springer}
}

@inproceedings{kading2016fine,
  title={Fine-tuning deep neural networks in continuous learning scenarios},
  author={K{\"a}ding, Christoph and Rodner, Erik and Freytag, Alexander and Denzler, Joachim},
  booktitle={Asian Conference on Computer Vision},
  pages={588--605},
  year={2016},
  organization={Springer}
}

@inproceedings{lin2014microsoft,
  title={Microsoft coco: Common objects in context},
  author={Lin, Tsung-Yi and Maire, Michael and Belongie, Serge and Hays, James and Perona, Pietro and Ramanan, Deva and Doll{\'a}r, Piotr and Zitnick, C Lawrence},
  booktitle={European conference on computer vision},
  pages={740--755},
  year={2014},
  organization={Springer}
}

@inproceedings{liu2022p,
  title={P-tuning: Prompt tuning can be comparable to fine-tuning across scales and tasks},
  author={Liu, Xiao and Ji, Kaixuan and Fu, Yicheng and Tam, Weng and Du, Zhengxiao and Yang, Zhilin and Tang, Jie},
  booktitle={Proceedings of the 60th Annual Meeting of the Association for Computational Linguistics (Volume 2: Short Papers)},
  pages={61--68},
  year={2022}
}

@inproceedings{nilsback2008automated,
  title={Automated flower classification over a large number of classes},
  author={Nilsback, Maria-Elena and Zisserman, Andrew},
  booktitle={2008 Sixth Indian conference on computer vision, graphics \& image processing},
  pages={722--729},
  year={2008},
  organization={IEEE}
}

@inproceedings{parkhi2012cats,
  title={Cats and dogs},
  author={Parkhi, Omkar M and Vedaldi, Andrea and Zisserman, Andrew and Jawahar, CV},
  booktitle={2012 IEEE conference on computer vision and pattern recognition},
  pages={3498--3505},
  year={2012},
  organization={IEEE}
}

@article{peters2019tune,
  title={To tune or not to tune? adapting pretrained representations to diverse tasks},
  author={Peters, Matthew E and Ruder, Sebastian and Smith, Noah A},
  journal={arXiv preprint arXiv:1903.05987},
  year={2019}
}

@article{si2024flora,
  title={Flora: Low-rank core space for n-dimension},
  author={Si, Chongjie and Wang, Xuehui and Yang, Xue and Xu, Zhengqin and Li, Qingyun and Dai, Jifeng and Qiao, Yu and Yang, Xiaokang and Shen, Wei},
  journal={arXiv preprint arXiv:2405.14739},
  volume={10},
  year={2024}
}

@inproceedings{sung2022vl,
  title={Vl-adapter: Parameter-efficient transfer learning for vision-and-language tasks},
  author={Sung, Yi-Lin and Cho, Jaemin and Bansal, Mohit},
  booktitle={Proceedings of the IEEE/CVF conference on computer vision and pattern recognition},
  pages={5227--5237},
  year={2022}
}

@inproceedings{szegedy2015going,
  title={Going deeper with convolutions},
  author={Szegedy, Christian and Liu, Wei and Jia, Yangqing and Sermanet, Pierre and Reed, Scott and Anguelov, Dragomir and Erhan, Dumitru and Vanhoucke, Vincent and Rabinovich, Andrew},
  booktitle={Proceedings of the IEEE conference on computer vision and pattern recognition},
  pages={1--9},
  year={2015}
}

@incollection{wang2022pre,
  title={Pre-training and fine-tuning},
  author={Wang, Jindong and Chen, Yiqiang},
  booktitle={Introduction to Transfer Learning: Algorithms and Practice},
  pages={125--140},
  year={2022},
  publisher={Springer}
}

@inproceedings{yin2024parameter,
  title={Parameter-efficient is not sufficient: Exploring parameter, memory, and time efficient adapter tuning for dense predictions},
  author={Yin, Dongshuo and Han, Xueting and Li, Bin and Feng, Hao and Bai, Jing},
  booktitle={Proceedings of the 32nd ACM International Conference on Multimedia},
  pages={1398--1406},
  year={2024}
}

@inproceedings{yin20231,
  title={1\% vs 100\%: Parameter-efficient low rank adapter for dense predictions},
  author={Yin, Dongshuo and Yang, Yiran and Wang, Zhechao and Yu, Hongfeng and Wei, Kaiwen and Sun, Xian},
  booktitle={Proceedings of the IEEE/CVF conference on computer vision and pattern recognition},
  pages={20116--20126},
  year={2023}
}

@inproceedings{zhou2017scene,
  title={Scene parsing through ade20k dataset},
  author={Zhou, Bolei and Zhao, Hang and Puig, Xavier and Fidler, Sanja and Barriuso, Adela and Torralba, Antonio},
  booktitle={Proceedings of the IEEE conference on computer vision and pattern recognition},
  pages={633--641},
  year={2017}
}

@article{yang2022focal,
  title={Focal modulation networks},
  author={Yang, Jianwei and Li, Chunyuan and Dai, Xiyang and Gao, Jianfeng},
  journal={Advances in Neural Information Processing Systems},
  volume={35},
  pages={4203--4217},
  year={2022}
}

@inproceedings{liu2022swin,
  title={Swin transformer v2: Scaling up capacity and resolution},
  author={Liu, Ze and Hu, Han and Lin, Yutong and Yao, Zhuliang and Xie, Zhenda and Wei, Yixuan and Ning, Jia and Cao, Yue and Zhang, Zheng and Dong, Li and others},
  booktitle={Proceedings of the IEEE/CVF conference on computer vision and pattern recognition},
  pages={12009--12019},
  year={2022}
}

@inproceedings{fan2021multiscale,
  title={Multiscale vision transformers},
  author={Fan, Haoqi and Xiong, Bo and Mangalam, Karttikeya and Li, Yanghao and Yan, Zhicheng and Malik, Jitendra and Feichtenhofer, Christoph},
  booktitle={Proceedings of the IEEE/CVF international conference on computer vision},
  pages={6824--6835},
  year={2021}
}

@inproceedings{li2022mvitv2,
  title={Mvitv2: Improved multiscale vision transformers for classification and detection},
  author={Li, Yanghao and Wu, Chao-Yuan and Fan, Haoqi and Mangalam, Karttikeya and Xiong, Bo and Malik, Jitendra and Feichtenhofer, Christoph},
  booktitle={Proceedings of the IEEE/CVF conference on computer vision and pattern recognition},
  pages={4804--4814},
  year={2022}
}

@article{vaswani2017attention,
  title={Attention is all you need},
  author={Vaswani, Ashish and Shazeer, Noam and Parmar, Niki and Uszkoreit, Jakob and Jones, Llion and Gomez, Aidan N and Kaiser, {\L}ukasz and Polosukhin, Illia},
  journal={Advances in neural information processing systems},
  volume={30},
  year={2017}
}

@inproceedings{sandler2018mobilenetv2,
  title={Mobilenetv2: Inverted residuals and linear bottlenecks},
  author={Sandler, Mark and Howard, Andrew and Zhu, Menglong and Zhmoginov, Andrey and Chen, Liang-Chieh},
  booktitle={Proceedings of the IEEE conference on computer vision and pattern recognition},
  pages={4510--4520},
  year={2018}
}

@article{hendrycks2016gaussian,
  title={Gaussian Error Linear Units (Gelus)},
  author={Hendrycks, Dan and Gimpel, Kevin},
  journal={arXiv preprint arXiv:1606.08415},
  year={2016}
}

@article{howard2017mobilenets,
  title={Mobilenets: Efficient convolutional neural networks for mobile vision applications},
  author={Howard, Andrew G and Zhu, Menglong and Chen, Bo and Kalenichenko, Dmitry and Wang, Weijun and Weyand, Tobias and Andreetto, Marco and Adam, Hartwig},
  journal={arXiv preprint arXiv:1704.04861},
  year={2017}
}

@inproceedings{lin2017feature,
  title={Feature pyramid networks for object detection},
  author={Lin, Tsung-Yi and Doll{\'a}r, Piotr and Girshick, Ross and He, Kaiming and Hariharan, Bharath and Belongie, Serge},
  booktitle={Proceedings of the IEEE conference on computer vision and pattern recognition},
  pages={2117--2125},
  year={2017}
}

@inproceedings{bossard2014food,
  title={Food-101--mining discriminative components with random forests},
  author={Bossard, Lukas and Guillaumin, Matthieu and Van Gool, Luc},
  booktitle={European conference on computer vision},
  pages={446--461},
  year={2014},
  organization={Springer}
}

@inproceedings{zhao2017pyramid,
  title={Pyramid scene parsing network},
  author={Zhao, Hengshuang and Shi, Jianping and Qi, Xiaojuan and Wang, Xiaogang and Jia, Jiaya},
  booktitle={Proceedings of the IEEE conference on computer vision and pattern recognition},
  pages={2881--2890},
  year={2017}
}

@article{chen2014semantic,
  title={Semantic image segmentation with deep convolutional nets and fully connected crfs},
  author={Chen, Liang-Chieh and Papandreou, George and Kokkinos, Iasonas and Murphy, Kevin and Yuille, Alan L},
  journal={arXiv preprint arXiv:1412.7062},
  year={2014}
}

@inproceedings{hu2018squeeze,
  title={Squeeze-and-excitation networks},
  author={Hu, Jie and Shen, Li and Sun, Gang},
  booktitle={Proceedings of the IEEE conference on computer vision and pattern recognition},
  pages={7132--7141},
  year={2018}
}

@article{graves2012long,
  title={Long short-term memory},
  author={Graves, Alex},
  journal={Supervised sequence labelling with recurrent neural networks},
  pages={37--45},
  year={2012},
  publisher={Springer}
}

@inproceedings{ioffe2015batch,
  title={Batch normalization: Accelerating deep network training by reducing internal covariate shift},
  author={Ioffe, Sergey and Szegedy, Christian},
  booktitle={International conference on machine learning},
  pages={448--456},
  year={2015},
  organization={pmlr}
}

@inproceedings{huang2016deep,
  title={Deep networks with stochastic depth},
  author={Huang, Gao and Sun, Yu and Liu, Zhuang and Sedra, Daniel and Weinberger, Kilian Q},
  booktitle={European conference on computer vision},
  pages={646--661},
  year={2016},
  organization={Springer}
}

@article{ba2016layer,
  title={Layer normalization},
  author={Ba, Jimmy Lei and Kiros, Jamie Ryan and Hinton, Geoffrey E},
  journal={arXiv preprint arXiv:1607.06450},
  year={2016}
}

@inproceedings{xiao2018unified,
  title={Unified perceptual parsing for scene understanding},
  author={Xiao, Tete and Liu, Yingcheng and Zhou, Bolei and Jiang, Yuning and Sun, Jian},
  booktitle={Proceedings of the European conference on computer vision (ECCV)},
  pages={418--434},
  year={2018}
}

@article{xin2024parameter,
  title        = {Parameter‑Efficient Fine‑Tuning for Pre‑Trained Vision Models: A Survey},
  author       = {Xin, Yi and Luo, Siqi and Zhou, Haodi and Du, Junlong and Liu, Xiaohong and Fan, Yue and Li, Qing and Du, Yuntao},
  journal      = {arXiv preprint arXiv:2402.02242},
  year         = {2024},
  note         = {v5},
  OPTurl          = {https://arxiv.org/abs/2402.02242}
}

@inproceedings{sahay2025mopeft,
  title        = {MoPEFT: A Mixture‑of‑PEFTs for the Segment Anything Model},
  author       = {Sahay, Rajat and Savakis, Andreas},
  booktitle    = {Proceedings of the IEEE/CVF Conference on Computer Vision and Pattern Recognition (CVPR) Workshops},
  year         = {2025},
  note         = {Workshop: Domain Generalization: Evolution, Breakthroughs, and Future Horizons},
  OPTurl          = {https://openaccess.thecvf.com/content/CVPR2025W/DG‑EBF/papers/Sahay_MoPEFT_A_Mixture‑of‑PEFTs_for_the_Segment_Anything_Model_CVPRW_2025_paper.pdf}
}
}

\end{document}